\documentclass[letterpaper, 10 pt, conference]{ieeeconf}
\IEEEoverridecommandlockouts
\usepackage{amsmath}
\usepackage{amssymb}
\usepackage{graphicx}
\usepackage{cite}
\usepackage{caption}
\usepackage{dblfloatfix}
\usepackage{xcolor}

\definecolor{nhred}{HTML}{D62727}
\definecolor{nhblue}{HTML}{1F77B4}
\definecolor{nhgray}{HTML}{807C7C}
\definecolor{nhgreen}{HTML}{4F7D75}

\newcommand\blfootnote[1]{%
  \begingroup
  \renewcommand\thefootnote{}\footnote{#1}%
  \addtocounter{footnote}{-1}%
  \endgroup
}
\providecommand{\mathcolor}[2]{{\color{#1}#2}}

\title{\LARGE \bf
Generalizable Robotic Insertion with World Models
}
\author{Nicklas Hansen\textcolor{black}{${}^{12\star}$},\hspace*{7pt}Iretiayo Akinola\textcolor{black}{${}^1$},\hspace*{7pt}Yijie Guo\textcolor{black}{${}^1$},\hspace*{7pt}Jie Xu\textcolor{black}{${}^1$},\hspace*{7pt}Bingjie Tang\textcolor{black}{${}^{13}$},\hspace*{7pt}\\
Hao Su\textcolor{black}{${}^2$},\hspace*{7pt}Xiaolong Wang\textcolor{black}{${}^2$},\hspace*{7pt}Abhishek Gupta\textcolor{black}{${}^1$},\hspace*{7pt}Dieter Fox\textcolor{black}{${}^1$},\hspace*{7pt}Yashraj Narang\textcolor{black}{${}^1$}
}

\begin{document}

\thispagestyle{empty}
\pagestyle{empty}

\twocolumn[{%
\renewcommand\twocolumn[1][]{#1}%
\maketitle
\vspace{-0.15in}
\begin{center}
    \centering
    \captionsetup{type=figure}
    \includegraphics[width=0.4\linewidth]{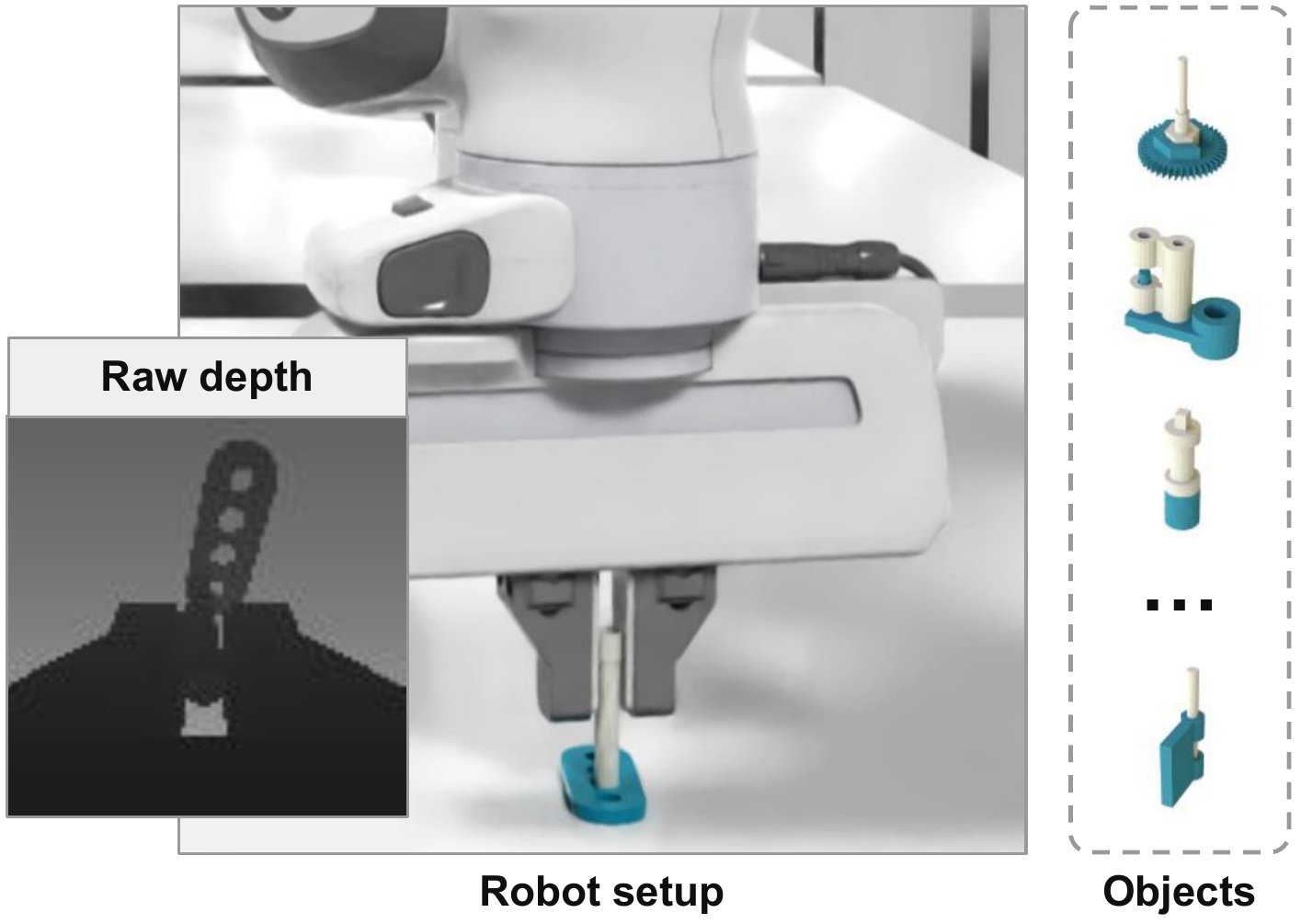}\hspace{0.25in}
    \includegraphics[width=0.26\linewidth]{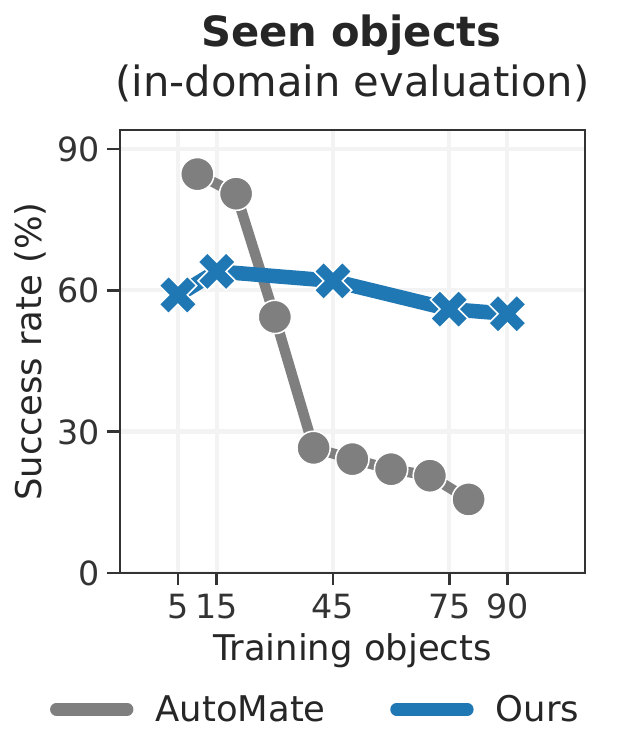}~
    \includegraphics[width=0.26\linewidth]{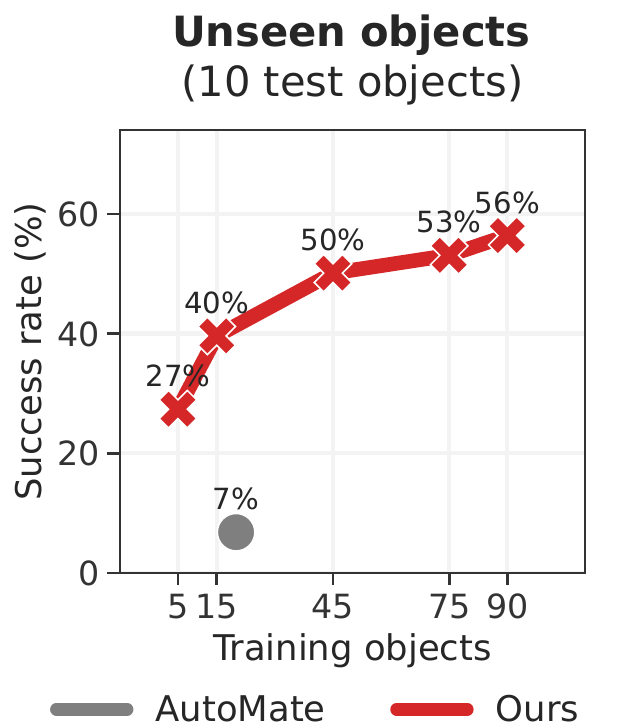}
    \captionof{figure}{\textbf{A generalist for robotic insertion.} When pretrained on a large number of objects (\textbf{{\textcolor{nhblue}{blue}}}), our method scales substantially better than the previous state-of-the-art generalist for insertion, \emph{AutoMate} \cite{tang2024automate} (\textbf{{\textcolor{nhgray}{gray}}}). Furthermore, our method generalizes \emph{zero-shot} to unseen object geometries (\textbf{\textcolor{nhred}{red}}), with generalization improving as number of training objects increases.}
    \label{fig:teaser}
\end{center}%
}]

\begin{abstract}
Robotic assembly in high-mixture settings requires adaptable systems that can handle diverse parts, yet current approaches typically rely on policies specialized to each insertion task. Although this can reach high success rates, it makes the process of deploying systems for new problems tedious and time consuming. We present a framework for generalizable insertion using world models that combine robot proprioceptive information with raw visual observations captured by a wrist-mounted camera. Our model-based approach trains a \emph{single} world model on up to 90 insertion tasks with geometrically diverse parts, achieving 56$\%$ zero-shot success on unseen objects with unknown geometry compared to just 7$\%$ with a model-free baseline. Importantly, performance improves as more objects are included in the training dataset, demonstrating strong scalability. Lastly, finetuning the generalist model on held-out objects significantly enhances data-efficiency compared to training from scratch and, in some cases, achieves better asymptotic performance. To our knowledge, this is the first system capable of assembling \emph{unseen} objects in an entirely data-driven manner, and thus represents a significant step toward scalable, generalizable robotic assembly systems.\blfootnote{${}^{\star}$Work completed during an internship at NVIDIA.}
\blfootnote{${}^{1}$NVIDIA. ${}^{2}$University of California San Diego. ${}^{3}$University of Southern California. Correspondence to Nicklas Hansen \texttt{<nihansen@ucsd.edu>}.}
\end{abstract}

\section{INTRODUCTION}
\label{sec:introduction}

Autonomous robotic assembly \cite{xu2019compare,jia2019threaded,Willis2022JoinABLe,tian2022assemble,tang2023industreal,tang2024automate,noseworthy2025forge} is a challenging yet important research topic in manufacturing and automation, where precise manipulation of diverse objects is needed. State-of-the-art approaches in industry typically rely on hand-crafted controllers
that cannot easily generalize across different object geometries and scene configurations. Existing learning-based systems in research predominantly use specialist policies, requiring extensive training (and often part-specific tuning) for each specific object pair, which fundamentally limits scalability and adaptability. The key challenge in going beyond specialist policies has been the inability of the currently-chosen algorithms to tractably obtain controllers that can achieve both dexterity and generalization.

In contrast to current systems, we envision future robotic assembly systems that have the flexibility and robustness required to manipulate new objects with minimal additional engineering, assumptions, or privileged information. To realize this vision, we argue that such a generalizable robotic assembly system demands adaptive control policies as well as robust visual perception capable of inferring detailed object geometry and poses directly from raw visual feedback within a closed-loop control system. While the current paradigms for acquiring learning-based controllers for assembly do not conceptually preclude such a vision, they struggle to scale up the learning of generalist behaviors to large numbers of tasks and rich perceptual inputs~\cite{tang2024automate}. This begs the question - as opposed to simply exposing methods to more data and compute, does a change in learning \emph{algorithm} potentially help develop generalist assembly policies more effectively?

Our research addresses these questions by introducing \emph{InsertionWM}: a learning-based framework that enables zero-shot insertion of unseen objects with entirely unknown object geometries, using visual world models. Concretely, we develop a scalable, yet data-efficient system for \emph{general-purpose} robotic assembly that leverages world models trained from raw visual observations captured by a wrist-mounted camera, as well as robot proprioceptive information. Departing from previous work that leverage model-free reinforcement learning (RL) for acquiring robotic assembly controllers \cite{tang2023industreal,tang2024automate,noseworthy2025forge}, we posit that model-based methods enable more efficient multi-task, vision-based policy learning. Intuitively, a predictive world model yields a dense, self-supervised signal from every transition -- decoupling representation learning from the sparse reward -- and supports planning at decision time; both are valuable when one high-capacity visual encoder is shared across many tasks, where model-free policy gradients are sample-inefficient and prone to task interference. Doing so allows for training generalist, vision-based policies across wider ranges of problems, without compromising dexterity, a challenge that has proven difficult with prior policy gradient-driven RL approaches to assembly.

Our approach, InsertionWM, learns a generalizable policy \emph{without} relying on object-specific priors. By training a multi-task policy on as many as $90$ diverse assemblies from AutoMate \cite{tang2024automate}, our method achieves up to $56\%$ \emph{zero-shot} success rate on unseen assemblies with \emph{unknown} geometry, and we show that performance improves as more assemblies are added, demonstrating strong scaling behavior; see Figure~\ref{fig:teaser} for a summary of results. Lastly, we find that finetuning a generalist model on held-out assemblies greatly improves data-efficiency compared to training from scratch and, in some cases, achieves better asymptotic performance.

Notably, the key contribution of this work is not to propose a new algorithm but rather to demonstrate that \textbf{\emph{visual model-based RL algorithms have immense untapped potential for industrial assembly}}, allowing for much broader and more performant generalist, multi-task policies than previous work. We carefully study the impact of various design choices for such a system, resulting in a practical guide to building general-purpose assembly systems using world models.

\section{RELATED WORK}
\label{sec:related-work}

Our work is at the intersection of robotic assembly and RL; this section summarizes prior work in these areas.

\textbf{Autonomous robotic assembly} tasks of interest to the research community often include mating of two distinct parts, with insertion of one object into another being a common yet difficult task due to intricate part geometries, asymmetry, precise control requirements, and the generally small scale of parts to be assembled \cite{gissler2019interlinked,xu2019compare,jia2019threaded, ferguson2021intersection, Willis2022JoinABLe,tian2022assemble}. Several recent works have explored learning of policies for robotic assembly via simulation environments designed to mimic a real-world setup, and these works demonstrate that such policies subsequently can be deployed on real hardware with limited reduction in performance \cite{fu20226d, tang2023industreal,tang2024automate,noseworthy2025forge}. However, due to the immense work required to build such simulation environments, development of better algorithms for policy learning has received comparably less attention from the research community. For example, AutoMate \cite{tang2024automate} proposes a system for training part-specific specialist policies to perform insertion in simulation which are then deployed on a real robot setup that closely matches the simulation environment. To do so, the authors curate and preprocess a set of 100 assembly asset pairs originally introduced by \cite{Willis2022JoinABLe,tian2022assemble}, and build a parallelizable simulation environment with observation and action spaces that are also accessible in the real world, as well as a shaped reward function that encourages reliable top-down insertion strategies. The authors demonstrate that their system enables training of Proximal Policy Optimization (PPO) \cite{schulman2017proximal} via online RL on individual object pairs, and include an exploratory experiment in which a number of learned specialist policies are distilled into a (blind) generalist insertion policy. In this work, we focus on the algorithmic aspect of robotic assembly and set out to develop a learning-based system that can insert \emph{unseen} objects with unknown geometry in a \emph{zero-shot} manner, relying purely on raw visual observations from a wrist-mounted camera.

\textbf{Reinforcement learning for robotics} has been of great interest to the research community over the past decade, with applications including tabletop manipulation \cite{kober2013rlsurvey, levine2016endtoend, pinto2017asymmetricactorcriticimagebased, nair2018visual, Zhang2018SOLARDS, zhan2020framework, lancaster2023modemv2, Feng2023Finetuning}, dexterous manipulation \cite{openai2019solving,zhu2019dexterous,qi2022hand,handa2023dextreme}, and outdoor locomotion \cite{lee2020quadrupedal, kumar2021rma, margolisyang2022rapid, cheng2023parkour, cheng2024express}. A common property of these applications is that they require learning robust control policies that can operate in increasingly noisy and unstructured environments (cannot easily be programmed), yet \emph{the tasks themselves are often quite forgiving in terms of precision of motions} \cite{elguea2023review}. In this work, we focus on learning \emph{precise} visual policies for assembly of (unseen) objects with unknown geometry, a significantly more challenging problem setting than what RL is typically used for in robotics literature. To achieve our goal, we build upon model-based RL algorithm TD-MPC2 \cite{Hansen2022tdmpc, hansen2024tdmpc2}, which has been shown empirically to outperform contemporary RL algorithms across a variety of continuous control tasks \cite{Hansen2025Newt}. Notably, this is (to the best of our knowledge) the first successful application of model-based RL algorithms in the area of robotic assembly.

\section{PRELIMINARIES}
\label{sec:preliminaries}

\textbf{Setup.} We develop a learning-based robotic assembly system capable of high-precision insertion of unseen objects with unknown geometry, entirely from raw visual observations captured by a wrist-mounted camera and robot proprioceptive information. To do so, we leverage Isaac \cite{makoviychuk2021isaac}, a GPU-accelerated and highly parallelizable simulation tool, for iterative data collection and RL. Concretely, our experimental setup consists of \emph{(1)} a Franka Panda manipulator with a parallel jaw gripper mounted to a table-top on which assembly occurs, \emph{(2)} a wrist-mounted camera mimicking the common Intel RealSense D435 RGB-D camera, and \emph{(3)} 200 simulated assets (100 object pairs) for training of continuous control policies. While our experiments are conducted in simulation, the hardware configuration mirrors platforms on which closely related work has been deployed in the real world \cite{fu20226d, tang2023industreal,tang2024automate,noseworthy2025forge}. We emphasize, however, that these prior demonstrations rely on low-dimensional pose estimation rather than raw visual inputs; transferring a vision-based world model such as ours therefore raises additional sim-to-real challenges discussed in Section~\ref{sec:conclusion}. Refer to Figure~\ref{fig:teaser} (\emph{left}) for an illustration of our setup.

\textbf{Robotic insertion with RL.} We model the robotic insertion problem as a Markov Decision Process (MDP) \cite{Bellman1957MDP} characterized by the tuple $(\mathcal{S}, \mathcal{A}, \mathcal{T}, \mathcal{R}, \gamma)$ where $\mathbf{s} \in \mathcal{S}$ are states, $\mathbf{a} \in \mathcal{A}$ are actions (6 DoF delta end-effector poses normalized to the $[-1, 1]$ interval), $\mathcal{T} \colon \mathcal{S} \times \mathcal{A} \mapsto \mathcal{S}$ is the environment transition (dynamics) function, $\mathcal{R} \colon \mathcal{S} \mapsto \mathbb{R}$ is a scalar reward function that encourages completion of the insertion task, and $\gamma$ is a constant discount factor. Since the full state of the environment $\mathbf{s}$ cannot be easily determined in real-world applications, we instead rely on raw depth observations $\mathbf{x} \in \mathbb{R}^{96\times 96}$ and robot proprioceptive information $\mathbf{q} \in \mathbb{R}^{24}$, and approximate $\mathbf{s}$ as $[\mathbf{x}, \mathbf{q}]$, resulting in a \emph{partially observable} MDP \cite{KAELBLING199899}. The objective is to learn a policy $\pi \colon \mathcal{S} \mapsto \mathcal{A}$ such that cumulative discounted rewards $\mathbb{E}_{\pi} \left[ \sum_{t=0}^{\infty} \gamma^{t} r_{t} \right]\,,~r_{t} \doteq \mathcal{R}(\mathbf{s}_{t})$ (denoted as \emph{return}) is maximized. In this work, we derive our policy $\pi$ by planning with a learned world model.

\textbf{Learning a world model with TD-MPC2} \cite{hansen2024tdmpc2}, a model-based RL algorithm that plans actions via local trajectory optimization in the latent space of a learned world model. Specifically, TD-MPC2 learns a latent world model from environment interaction data and leverages Model Predictive Path Integral (MPPI) \cite{williams2015model} with a learned policy prior as a sampling-based trajectory optimizer during inference (\emph{planning}). All components of the world model are learned end-to-end from data using an optimization objective that combines auto-regressive joint-embedding prediction \cite{Grill2020BootstrapYO}, reward prediction, and a temporal difference (TD) \cite{Sutton1988LearningTP,Lillicrap2016ContinuousCW} loss, \emph{without} decoding raw observations. In this work, we build upon TD-MPC2 due to its demonstrably strong empirical performance on simpler robotic manipulation tasks, and develop a system for generalizable, vision-based robotic insertion. In the following, we provide more details on the TD-MPC2 algorithm as used in our work.\vspace{0.05in}\\
~$\boldsymbol{-}$~\textbf{\emph{Model architecture.}} TD-MPC2 learns a latent world model that consists of 5 components:\vspace{-0.05in}
\begin{equation}
    \label{eq:tdmpc-components}
    \begin{array}{lll}
        \text{Encoder} & \mathbf{z} = h(\mathbf{s}) & \color{nhgray}{\vartriangleright\text{Produce latent state}}\\
        \text{Dynamics} & \mathbf{z}' = d(\mathbf{z}, \mathbf{a}) & \color{nhgray}{\vartriangleright\text{Predict next latent state}}\\
        \text{Reward} & \hat{r} = R(\mathbf{z}) & \color{nhgray}{\vartriangleright\text{Predict reward $r$}}\\
        \text{Terminal value} & \hat{q} = Q(\mathbf{z}, \mathbf{a}) & \color{nhgray}{\vartriangleright\text{Predict future return}}\\
        \text{Policy prior} & \hat{\mathbf{a}} \sim p(\mathbf{z}) & \color{nhgray}{\vartriangleright\text{Predict action}}
    \end{array}
\end{equation}
where $\mathbf{z}$ is the latent state. During training, our agent autonomously collects data via interaction, and selects actions via planning. Our agent maintains a limited capacity (first-in-first-out) dataset (\emph{replay buffer}) $\mathcal{B}$ of collected trajectories, which is used to optimize the world model.\vspace{0.05in}\\
~$\boldsymbol{-}$~\textbf{\emph{Training objective.}} The $h,d,R,Q$ components of the world model are jointly optimized to minimize the objective
\begin{equation}
    \mathcal{L}\left(\theta\right) \doteq \mathop{\mathbb{E}}_{\left(\mathbf{s}, \mathbf{a}, r, \mathbf{s}'\right)_{0:H} \sim \mathcal{B}} \left[ \sum_{t=0}^{H} \lambda^{t} L(\theta; t) \right]\,,
\end{equation}
where $\lambda \in \mathbb{R}_{+}$ is a constant coefficient that weighs near-term predictions higher than predictions further into the future, $H$ is a fixed sequence length (horizon), and $L(\theta; t)$ is the single-step loss\vspace{-0.05in}
\begin{align}
    \label{eq:loss-terms}
    L(\theta; t) &= \mathcolor{nhblue}{\underbrace{\mathcolor{black}{\operatorname{CE}(\hat{r}_{t}, r_{t})}}_{\text{Reward prediction}}} + \mathcolor{nhblue}{\underbrace{\mathcolor{black}{\operatorname{CE}(\hat{q}_{t}, q_{t})}}_{\text{Value prediction}}}\\&+ \mathcolor{nhblue}{\underbrace{\mathcolor{black}{\|\ \mathbf{z}_{t}' - \operatorname{sg}(h(\mathbf{s}_{t}')) \|^{2}_{2}}}_{\text{Joint-embedding prediction}}}\,.
\end{align}
Here, we use two-hot discretization of reward/value targets $(r_{t}, q_{t})$ as in TD-MPC2 and optimize cross-entropy ($\operatorname{CE}$) over bins, and $\operatorname{sg}$ is a stop-grad operator that prevents gradients from flowing back through latent state prediction targets. The policy prior is trained to maximize entropy and terminal value $Q(\mathbf{z}, \mathbf{a})$ at each step. Refer to \cite{Hansen2022tdmpc, hansen2024tdmpc2} for more details on the TD-MPC2 algorithm and its application to general RL problems.

\section{DATASET AND ENVIRONMENT}
\label{sec:dataset-environment}

\begin{figure}
    \centering
    \includegraphics[width=\linewidth]{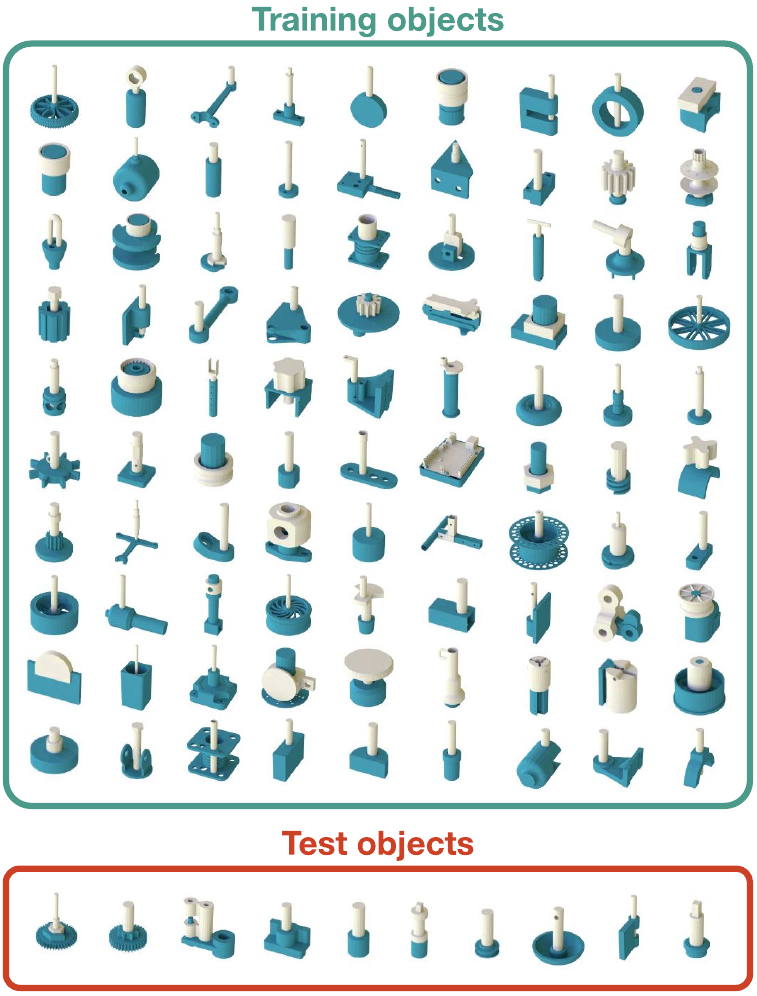}
    \vspace{-0.2in}
    \caption{\textbf{Assemblies.} We use 100 assemblies from \cite{tang2024automate}, which vary greatly in geometry. We reserve 10 assemblies as a held-out test set and train on the remaining 90. Training and test assemblies are drawn from the same data distribution.}
    \label{fig:automate-objects}
\end{figure}

\begin{figure*}[t]
    \centering
    \includegraphics[width=\linewidth]{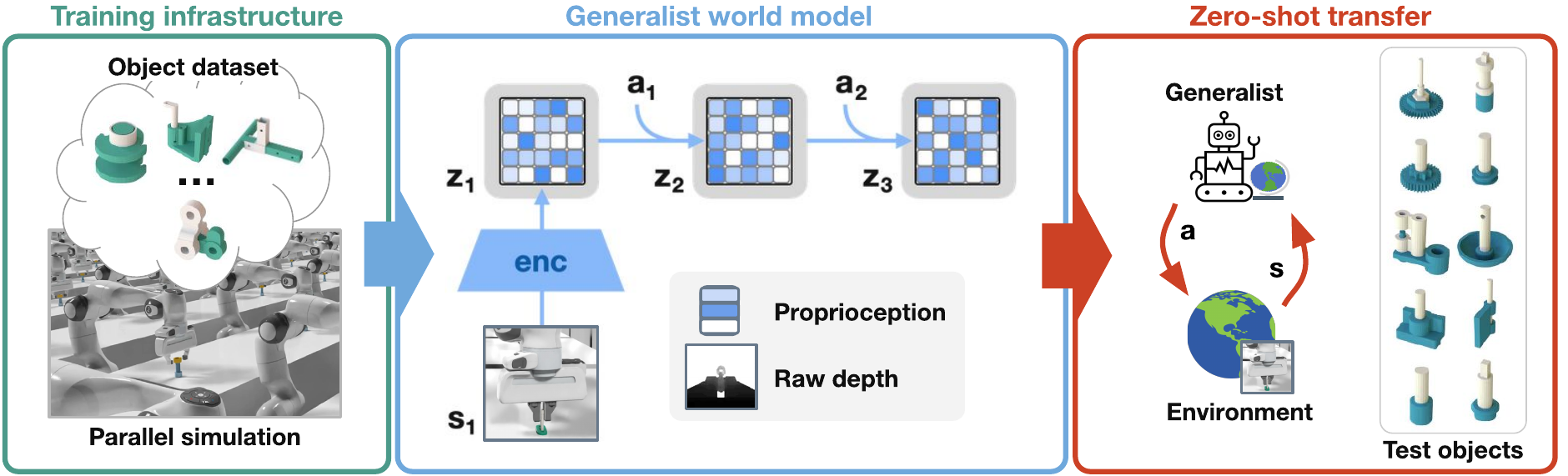}
    \vspace{-0.2in}
    \caption{\textbf{Overview.} We present a learning-based system for \textcolor{nhred}{\textbf{zero-shot}} robotic insertion of unseen objects with unknown geometry. Our framework, \emph{InsertionWM}, learns a \textcolor{nhblue}{\textbf{generalizable world model}} that takes robot proprioceptive information and raw depth observations captured by a wrist-mounted camera as input, and outputs delta end-effector pose commands.}
    \label{fig:method}
    \vspace{-0.15in}
\end{figure*}

\textbf{Dataset for robotic assembly.} We base our experiments on the assembly asset dataset proposed by \cite{tang2024automate}, which consists of $100$ geometrically diverse two-part assemblies that themselves are based on previous work \cite{Willis2022JoinABLe, tian2022assemble}. Despite significant progress in simulation tools in recent years, making assembly assets simulatable remains a tedious, human labor-intensive task, and thus we limit ourselves to these $100$ object pairs. We emphasize, however, that our method is not limited to a fixed number of objects and can -- without any algorithmic changes -- be applied to larger assembly datasets as they become available. The assemblies considered in this work are geometrically diverse and often feature asymmetries along at least one axis. As a result, insertion strategy can vary significantly between objects, necessitating the development of geometry-aware policies. Figure~\ref{fig:automate-objects} provides an overview of our assets; the \textbf{\textcolor{nhgray}{\emph{plug}}} is to be inserted into the \textbf{\textcolor{nhgreen}{\emph{socket}}}.

\textbf{Simulation environment.} We build upon the environment proposed by AutoMate \cite{tang2024automate}, which features a Franka Panda manipulator equipped with a parallel-jaw gripper, and closely follow their experimental setup to ensure that comparison to prior work is fair. The socket is randomly initialized within reach of the robot, in an upright pose such that an approximately top-down insertion is possible. The plug is similarly randomly initialized and is grasped via motion planning before the start of each RL trajectory. To facilitate learning and improve rate of convergence, we employ a simple training curriculum that adjusts the initial state distribution based on current training success rate. The curriculum is adjusted to each assembly individually when training multi-assembly models. Observations and rewards are defined as follows:\vspace{0.05in}\\
$\boldsymbol{-}$ \textbf{\emph{Observations.}} We consider observations that include raw depth observations $\mathbf{x} \in \mathbb{R}^{96\times 96}$ from a wrist-mounted camera, as well as robot proprioceptive information $\mathbf{q} \in \mathbb{R}^{24}$. The proprioceptive vector consists of robot joint angles ($\mathbb{R}^{7}$), noisy estimates of the 6D end-effector and socket poses ($2 \times \mathbb{R}^{7}$, where each $\mathbb{R}^{7}$ comprises a 3D position and a 4D unit quaternion) with white noise applied to mimic real-world sensor noise, and a goal position ($\mathbb{R}^{3}$) at which the plug is considered fully inserted. While prior work has considered strictly low-dimensional state observations, we find that visual observations can \emph{significantly} improve insertion of small, thin objects.\vspace{0.05in}\\
$\boldsymbol{-}$ \textbf{\emph{Reward function.}} We use a reward function
\begin{equation}
    \mathcal{R}(\mathbf{s}, \mathbf{a}) \doteq r_{\textnormal{penetration}} ( r_{\textnormal{dist}} + r_{\textnormal{imitation}} + r_{\textnormal{success}} )
\end{equation}
where $r_{\textnormal{penetration}}$ is a scaling penalty for interpenetration errors in the rare event that such simulation errors occur due to \emph{e.g.} very small or thin meshes, $r_{\textnormal{dist}}$ is the negative distance between plug and goal, $r_{\textnormal{imitation}}$ is an imitation-based reward that encourages the policy to mimic a small set of demonstrations derived by reversing procedurally-generated disassembly trajectories \cite{Peng2018Deep, tang2024automate}, and $r_{\textnormal{success}}$ is a binary term that rewards task success. We omit constant coefficients that balance each term for clarity. This reward function was originally proposed by \cite{tang2023industreal} and later adopted by \cite{tang2024automate}; we use the reward function without modification and merely provide this description for completeness.

\section{GENERALIZABLE ROBOTIC INSERTION}
\label{sec:method}

We present \emph{InsertionWM}: a learning-based system for \emph{zero-shot robotic insertion of unseen objects with unknown geometry}. Our framework learns a generalizable world model that at each time step takes robot proprioceptive information and raw depth observations captured by a wrist-mounted camera as input, and outputs delta end-effector pose commands for the robot to execute via planning with the learned world model. Crucially, InsertionWM does not rely on privileged information such as CAD models or object IDs during deployment, making deployment on real hardware feasible in the future. Figure~\ref{fig:method} provides an overview of our proposed system. In this section, we describe algorithmic details specific to our problem setting.

\textbf{Encoder for multi-modal observations.} Previous work on learning-based robotic assembly \cite{tang2023industreal, tang2024automate, noseworthy2025forge} focus on learning object-specific (specialist) policies for each unique assembly, which does not require substantial knowledge of object geometry beyond what can be deduced through robot proprioceptive information. However, we argue that visual feedback is necessary if we are to learn a \emph{single} policy (or world model) that can manipulate \emph{multiple} objects, including objects not seen during training. To this end, we additionally provide our world model with raw depth information from a wrist-mounted camera, and design the encoder $h$ to consist of three modules: \emph{(1)} a shallow ConvNet for depth observations, \emph{(2)} a 2-layer MLP for robot proprioceptive information, and \emph{(3)} a 2-layer MLP that fuses the two encoder outputs into a single latent state $\mathbf{z}$, formally defined as\vspace{-0.05in}
\begin{equation}
    \label{eq:encoder}
    h(\mathbf{x}, \mathbf{q}) \doteq h_{\textnormal{fuse}} \left( h_{\textnormal{depth}}(\mathbf{x}) + h_{\textnormal{prop}}(\mathbf{q}) \right)\,,
\end{equation}
where each of $h_{\textnormal{fuse}}, h_{\textnormal{depth}}, h_{\textnormal{prop}}$ apply SimNorm \cite{hansen2024tdmpc2} normalization to their output. We use multi-modal observations for both our single-object and multi-object world models. As our experimental results will reveal, the addition of visual observations is crucial to the performance of our framework.

\begin{figure}
    \centering
    \includegraphics[width=.1625\linewidth]{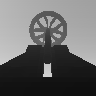}\hspace{0.1em}%
    \includegraphics[width=.1625\linewidth]{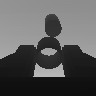}\hspace{0.1em}%
    \includegraphics[width=.1625\linewidth]{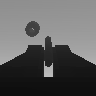}\hspace{0.1em}%
    \includegraphics[width=.1625\linewidth]{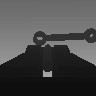}\hspace{0.1em}%
    \includegraphics[width=.1625\linewidth]{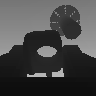}\hspace{0.1em}%
    \includegraphics[width=.1625\linewidth]{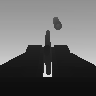}\\[0.1em]
    \includegraphics[width=.1625\linewidth]{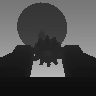}\hspace{0.1em}%
    \includegraphics[width=.1625\linewidth]{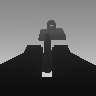}\hspace{0.1em}%
    \includegraphics[width=.1625\linewidth]{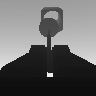}\hspace{0.1em}%
    \includegraphics[width=.1625\linewidth]{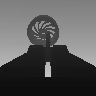}\hspace{0.1em}%
    \includegraphics[width=.1625\linewidth]{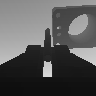}\hspace{0.1em}%
    \includegraphics[width=.1625\linewidth]{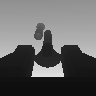}%
    \caption{\textbf{Visual observations.} A wrist-mounted depth camera provides rich visual information, allowing our model-based method to infer object geometry in a data-driven manner.}
    \label{fig:visual-observations}
    \vspace{-0.15in}
\end{figure}

\textbf{Zero-shot generalization.} We train both single-object insertion specialists and generalizable multi-object insertion policies via environment interaction. To fully leverage the parallelizable nature of our simulation environment, we choose to train our agents across multiple robotic insertion instances simulated in parallel. When training specialists, we randomly initialize the same assembly with different initial scene configurations across each environment instance, while we maintain one environment instance per unique assembly when training multi-task world models, \emph{i.e.}, we simulate $90$ environment instances each with one of $90$ unique assemblies from our dataset, with significant randomization in scene configuration. This guarantees a uniform distribution across assemblies, as well as robustness to object position and orientation. While providing the model with \emph{e.g.} a one-hot encoding of object IDs could potentially improve performance on \emph{known} objects in a multi-task setting, such a representation cannot easily be transferred zero-shot to new objects. Thus, \emph{we opt to rely solely on visual information for object disambiguation}. Figure~\ref{fig:visual-observations} shows visual observations captured by our wrist-mounted camera.

\section{EXPERIMENTS}
\label{sec:experiments}
\vspace{-0.05in}
We validate our approach, \emph{InsertionWM}, through rigorous experimental evaluation in the setup described in Section~\ref{sec:dataset-environment}, including evaluation of both single-assembly \emph{specialist} policies as well as multi-assembly \emph{generalist} policies, with our key performance metric being the average insertion success rate across all available assets.

\subsection{Experimental Details}
\label{sec:experiments-details}

\textbf{Evaluation.} All assemblies introduced in \cite{tang2024automate} have uniquely numbered identifiers. We reserve the last $10$ of $100$ assemblies (according to their IDs) as a fixed held-out \emph{test} set with geometries that are still in-distribution but not encountered during training. train generalist world models on the remaining $90$ objects. All reported \emph{test} rates refer to these held-out assemblies. When training generalists on only a subset of these objects, we simply select subsets according to their IDs in ascending order. This simple training and test set procedure serves to mitigate any selection bias that might occur from a more manual selection. Generalist success rates are averaged over all assemblies in the training (in-domain) or test (unseen) sets, respectively, and across $100$ trials per assembly. Specialist policies are trained independently on each of $100$ assemblies, and we report the mean success rate across $5$ independent training runs (random seeds) as well as $100$ trials per run.

\begin{figure*}
    \centering
    \includegraphics[width=\linewidth]{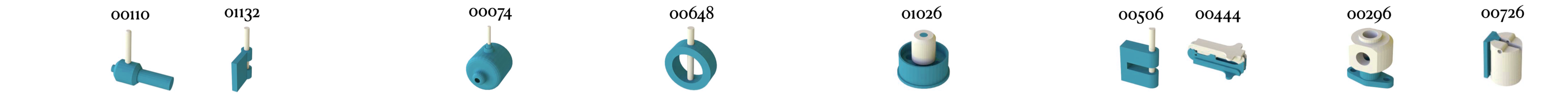}\\
    \includegraphics[width=\linewidth]{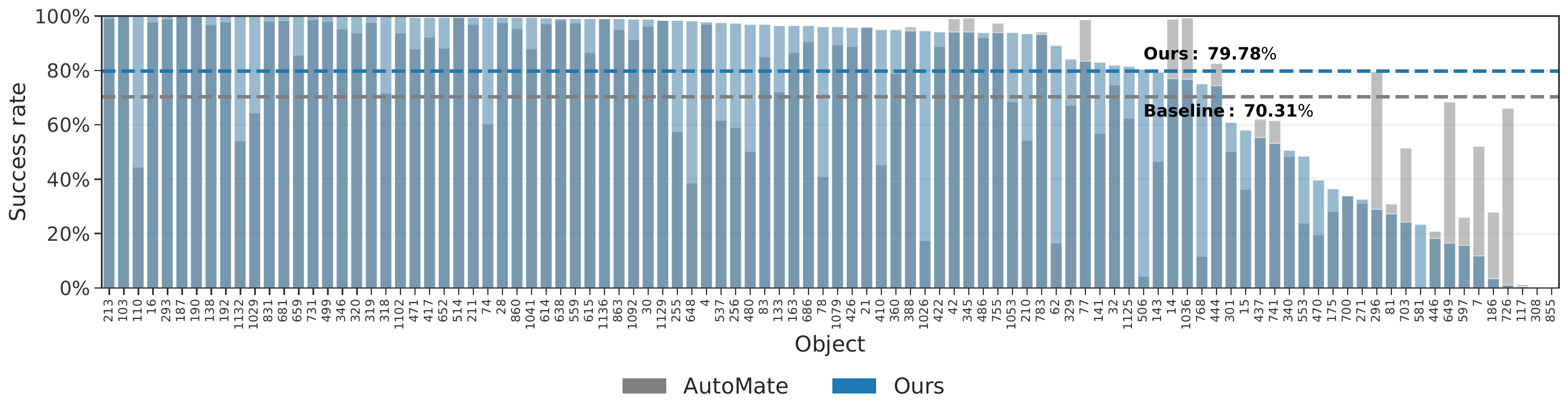}
    \vspace{-0.275in}
    \caption{\textbf{Specialist policies.} Our method (\textbf{\textcolor{nhblue}{blue}}) achieves strong single-assembly performance across 100 diverse object geometries ($79.78$\% overall success rate), exceeding AutoMate's (\textbf{\textcolor{nhgray}{gray}}) success rate of $70.31$\%. Assemblies are ordered by success rate; mean of 5 runs per assembly. Objects that result in higher levels of occlusion tend to have lower success rates.}
    \label{fig:auto-spec}
    \vspace{-0.1in}
\end{figure*}

\textbf{Baselines.} Our primary point of comparison is \textbf{AutoMate} \cite{tang2024automate}, the current state-of-the-art method for robotic insertion with RL. AutoMate trains insertion policies in simulation with PPO \cite{schulman2017proximal}, focusing mostly on specialist (single-assembly) policies with proprioceptive information as input. The trained specialist policies are blind (\emph{i.e.}, no visual information) and are not explicitly provided information about object geometry. In addition to these specialist policies, AutoMate also includes exploratory experiments on a generalist policy. This generalist policy is obtained by first training single-assembly specialist policies, and subsequently distilling them into a single multi-assembly policy using a combination of supervised behavioral cloning, DAgger \cite{ross2011reduction}, and online RL finetuning, while conditioning the policy on point cloud embeddings obtained from CAD models of each assembly. This contrasts with our generalist model that relies \emph{solely} on raw visual inputs and is trained purely with online RL. Our approach greatly simplifies the \emph{deployment} pipeline relative to AutoMate: although both use CAD models to build the training simulation, AutoMate further requires per-object CAD at deployment (point cloud embeddings for its generalist and a CAD-based pose-estimation pipeline), whereas our policy acts directly from raw visual observations and needs no object geometry (such as CAD models) at deployment. All AutoMate results are reproduced using the official codebase. Additionally, \textbf{we compare against a number of variations of our framework}, in particular \emph{(1)} \emph{"blind"} world models without visual inputs, \emph{(2)} world models with an alternative wrist camera placement, \emph{(3)} world models trained with and without a curriculum that adaptively adjusts initial conditions to be farther or closer to insertion success depending on current training success rate (following the curriculum setting in AutoMate), and \emph{(4)} a generalist with access to $100$ expert rollouts (generated by our specialist policies) per assembly. Naturally, access to such expert rollouts during training hinges on the availability of pretrained expert policies, thus severely limiting the scalability of the generalist training pipeline to larger sets of assemblies. The default formulation of our framework thus leverages a training curriculum rather than expert rollouts, and we merely include this additional baseline for completeness.

\textbf{Implementation details.} All world models considered in this work have $5$M learnable parameters. Observations include raw depth observations $\mathbf{x} \in \mathbb{R}^{96\times 96}$ from a wrist-mounted camera and robot proprioceptive information $\mathbf{q} \in \mathbb{R}^{24}$, and actions are $6$ DoF delta end-effector poses normalized to $[-1, 1]$. \emph{Specialist} world models are trained for $2$M steps vs. $50$M steps for AutoMate, and our \emph{generalists} are trained for $7$M steps. Training a specialist with visual inputs takes $40$h on a NVIDIA RTX 3090 GPU, and training a $90$-object generalist takes $3$ days on the same hardware.

\subsection{Results}
\label{sec:experiments-results}

\textbf{Single-assembly specialists.} We evaluate specialist policies across all $100$ assemblies, and report both individual and aggregate performance metrics for our method and AutoMate in Figure~\ref{fig:auto-spec}. For visual clarity, we order objects from left to right along the horizontal axis according to the average success rate achieved by our approach. Our approach achieves \textbf{$79.78\%$} success across all $100$ assemblies vs. $70.31\%$ for AutoMate\footnote{Reproduced numbers have been verified via correspondence with the AutoMate authors; they acknowledge a slight discrepancy in performance between their paper and released codebase.}. While such a substantial performance improvement is to be celebrated, we do observe a decrease in success rates across a few assemblies, \emph{e.g.} \texttt{296} and \texttt{726} (visualized in Figure~\ref{fig:auto-spec}). We conjecture that this may be due to differences in exploration between PPO (AutoMate) and TD-MPC2 (ours), as well as visual occlusion. Visual inspection of objects for which there is a large performance gap between our method and AutoMate -- positive or negative gain -- indicates that our approach tends to struggle with large sockets, presumably due to visual occlusion, while our method on the other hand performs significantly better than AutoMate on small, thin sockets. Regardless, these results clearly demonstrate the efficacy of our approach even in a single-assembly case.

\begin{figure*}[t]
    \centering
    \includegraphics[width=0.95\linewidth]{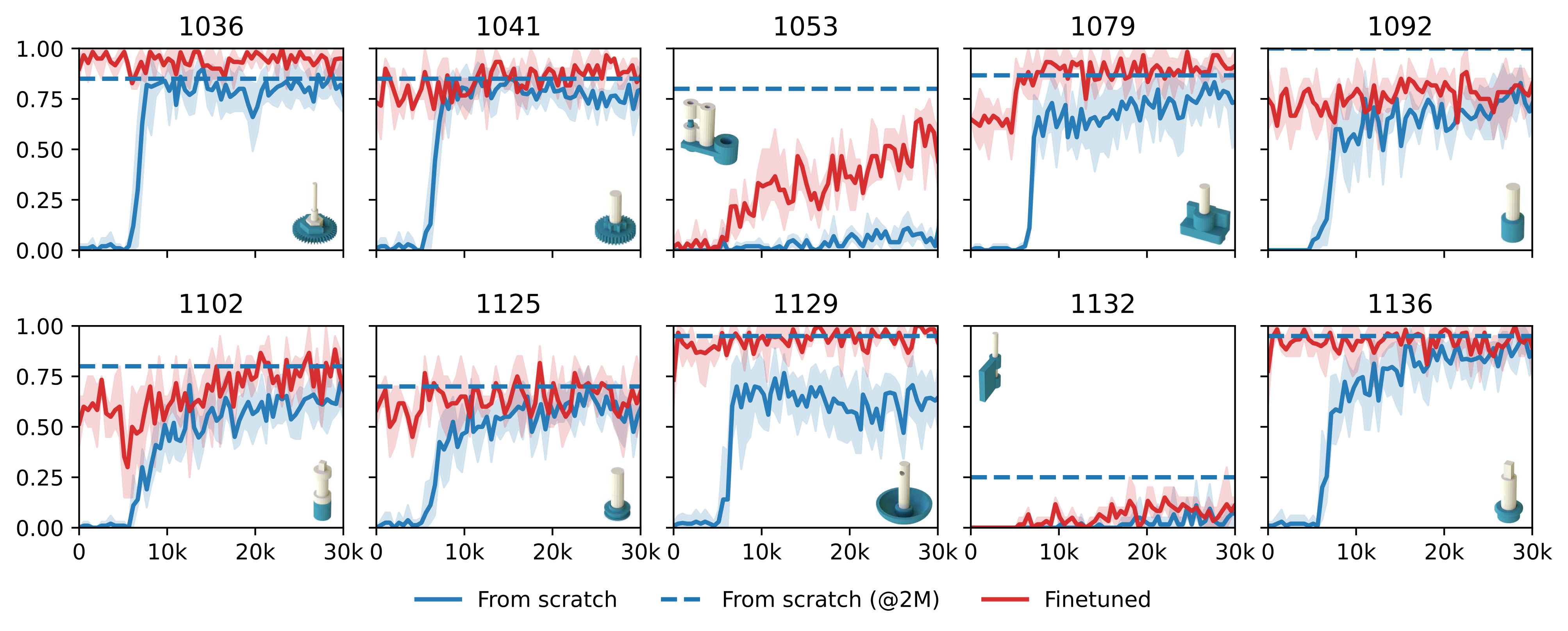}
    \vspace{-0.1in}
    \caption{\textbf{Finetuning.} Success rate vs. environment steps for two variants of our method: \emph{(1)} specialists learned from scratch, and \emph{(2)} our 90-object generalist finetuned on each of 10 held-out assemblies. The unique ID of each assembly is reported above each subplot. Finetuning greatly improves data-efficiency. Average of 5 random seeds; shaded area denotes $95\%$ CIs.}
    \label{fig:ft-curves}
    \vspace{-0.05in}
\end{figure*}

\begin{figure*}[t]
    \centering
    \includegraphics[width=0.24\textwidth]{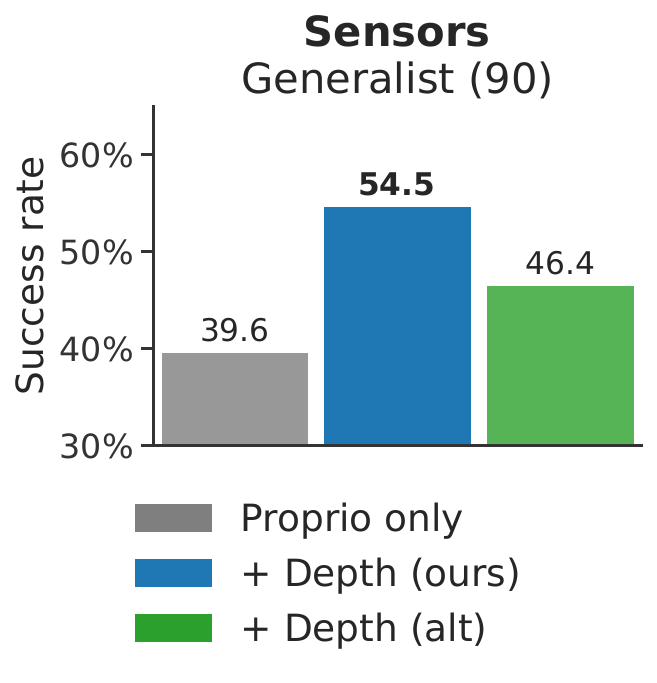}~
    \includegraphics[width=0.24\textwidth]{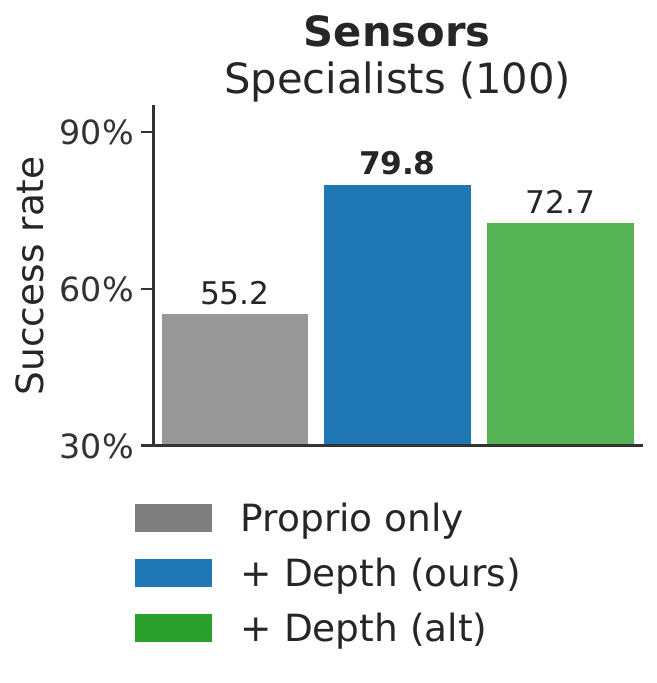}~
    \includegraphics[width=0.24\textwidth]{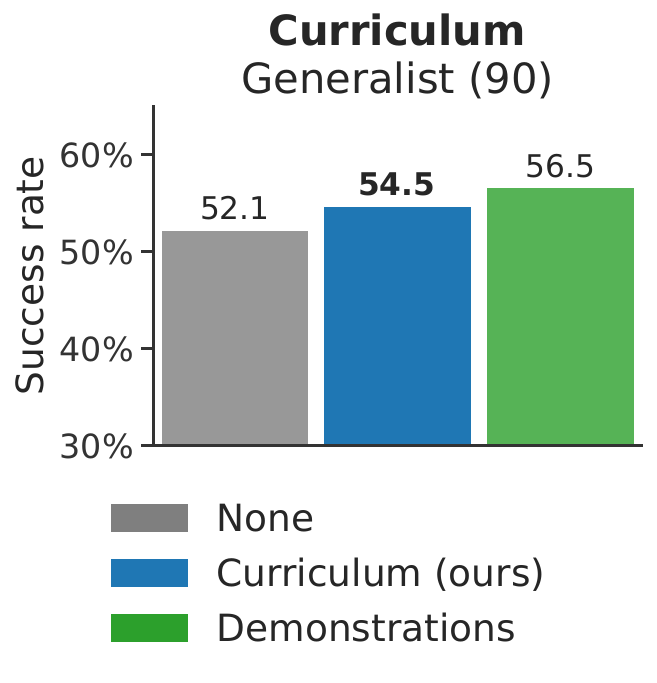}~
    \includegraphics[width=0.24\textwidth]{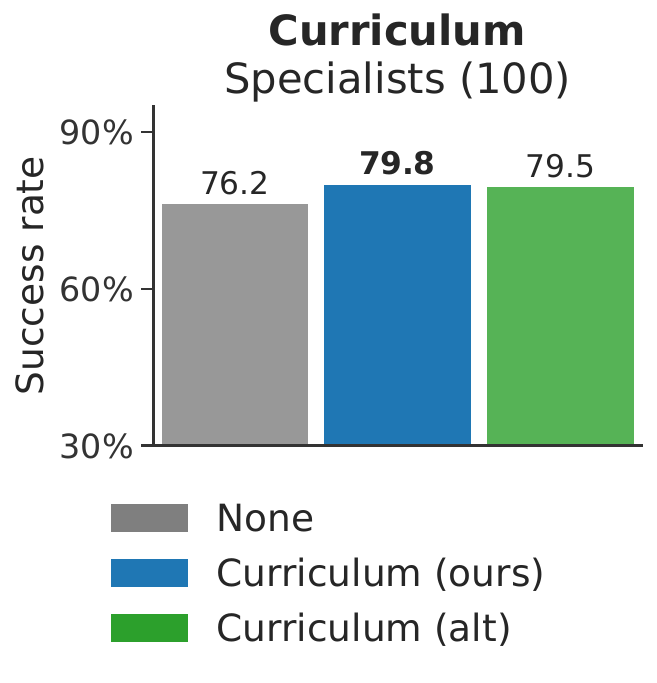}
    \vspace{-0.225in}
    \caption{\textbf{Ablations.} Success rate on in-domain objects (generalists trained on 90 objects), or averaged across all single-object specialist policies (trained on each of 100 objects). Our ablations highlight the relative importance of each design choice; \textbf{\textcolor{nhblue}{blue}} is the default formulation of our method. Visual observations are critical to both generalist and specialist world models.}
    \label{fig:ablations}
    \vspace{-0.125in}
\end{figure*}

\textbf{Multi-assembly generalists.} Our generalist results are shown in Figure~\ref{fig:teaser} (\emph{right}); we report numbers for both in-domain evaluations (\emph{seen} objects; left) and held-out evaluations (\emph{unseen} objects; right), as well as reference AutoMate numbers obtained from \cite{tang2024automate}. We note that the exact training and test splits for AutoMate are not known, but that our experimental setups are nonetheless comparable and that the difference in performance cannot be explained by a difference in training/test splits. It is also worth noting that the AutoMate generalist assumes access to point cloud embeddings of ground-truth CAD models for each assembly (including test assemblies), whereas our method relies solely on raw visual observations. Our results demonstrate that our model-based RL approach achieves considerably better scaling wrt. number of assemblies, achieving a $55.0\%$ success rate on $90$ seen assemblies compared to just $15.6\%$ for AutoMate on $80$ seen assemblies. Additionally, we find that the zero-shot generalization to unseen assemblies increases with the number of training assemblies, achieving up to $56\%$ \textbf{\textit{zero-shot}} success rate when trained on $90$ assemblies. This is a significant achievement as zero-shot performance on out-of-domain assemblies reaches in-domain performance at this scale. Based on our observed scaling trend, we hypothesize that in-domain and out-of-domain performances will saturate around this success rate as more assemblies become available in the future, assuming a similar data distribution and difficulty.

\textbf{Finetuning a generalist on held-out assemblies.} In other areas of robotics and artificial intelligence, one of the most useful capabilities of a generalist is to serve as a pretrained model that can be efficiently finetuned on problem-specific data. This may especially be true in the context of robotic assembly where assembly-specific data is limited. To further demonstrate the value of our approach, we additionally finetune our $90$-assembly generalist to each of the $10$ held-out assemblies; results are shown in Figure~\ref{fig:ft-curves}. We observe that starting from a generalist model greatly improves data-efficiency on new assemblies and, in some cases, exceeds the asymptotic performance of a specialist learned from scratch (objects \texttt{1036} and \texttt{1041}).

\textbf{Analysis \& ablations.} We conduct a series of ablations in both specialist (trained on each of $100$ objects) and generalist ($90$ training objects) and report results in Figure~\ref{fig:ablations}. First, we investigate the importance of vision by comparing our approach with \emph{(i)} a "blind" version of our method with access to proprioceptive information only, and \emph{(ii)} an alternative placement of the wrist-mounted camera which provides a better view of the grasped plug but is more prone to visual occlusion. Our results indicate that vision is indeed critical to performance, and our ablation on camera placement corroborates our earlier observation that our approach is prone to failure when the plug is large and (partially) occludes the socket. Our second ablation, shown in Figure~\ref{fig:ablations} (bottom), quantifies the effect of a curriculum that dynamically adapts initial conditions to the current per-assembly success rate. For completeness, we evaluate two such curricula: \emph{(i)} our default curriculum which starts at the easiest initial state distribution and gradually increases difficulty as success rate increases, and \emph{(ii)} an inverse curriculum that starts at the \emph{hardest} setting. Our results indicate that a curriculum consistently improves specialist and generalist performance when evaluated on the full state distribution, while the particular curriculum is less important. We also compare to a privileged version of our generalist that has access to $100$ expert rollouts (demonstrations) per assembly; we observe that this improves performance marginally but do no adopt this approach for our proposed generalist training pipeline as it scales poorly with larger object datasets due to the assumption of trained experts.

\section{CONCLUSION}
\label{sec:conclusion}
We present a learning-based system for \emph{zero-shot robotic insertion of unseen objects with unknown geometry}. Our approach achieves substantially better scaling than previous work, with task success rates improving as training data increases. Notably, it achieves an average $56\%$ zero-shot success rate on held-out tasks when trained on $90$ insertion tasks (objects). However, several opportunities for improvement remain; some are discussed in Section~\ref{sec:experiments-results} and we further emphasize the following ones: \emph{(i)} we observe that our system performs poorly on assemblies with significant visual occlusion, \emph{(ii)} we observe a slight decline in in-domain performance as number of training objects increases, \emph{(iii)} there is currently limited availability of assembly datasets for training generalists, and \emph{(iv)} deployment of visual policies on real hardware is still challenging due to potential simulation-to-real discrepancies, inaccurate depth sensing for small objects, and the need for real-time processing of visual inputs. Unlike prior systems that transfer to hardware via low-dimensional pose estimation and model-free policies \cite{tang2023industreal, tang2024automate, noseworthy2025forge}, a \emph{vision-based, model-based} approach such as ours faces added hurdles: commodity depth is noisy with blurred edges on small parts, and contact is hard to simulate faithfully. Although we already randomize initial poses, broader pose generalization and active perception that reduces geometric uncertainty before acting (\emph{e.g.} Fabrica \cite{tian2025fabrica}) are promising directions. We believe that future research in any of the aforementioned directions has immense potential in the area of robotic assembly.

\bibliographystyle{IEEEtran}
\bibliography{main}

\end{document}